# New Feature Detection Mechanism for Extended Kalman Filter Based Monocular SLAM with 1-Point RANSAC

Agniva Sengupta[(✉)] and Shafeeq Elanattil

Kritikal Solutions Pvt. Ltd., Bangalore, India
{agniva.sengupta,shafeeq.elanattil}@kritikalsolutions.com,
{i.agniva,eshafeeqe}@gmail.com
http://www.kritikalsolutions.com/

**Abstract.** We present a different approach of feature point detection for improving the accuracy of SLAM using single, monocular camera. Traditionally, Harris Corner detection, SURF or FAST corner detectors are used for finding feature points of interest in the image. We replace this with another approach, which involves building non-linear scale space representation of images using Perona and Malik Diffusion equation and computing the scale normalized Hessian at multiple scale levels (KAZE feature). The feature points so detected are used to estimate the state and pose of a mono camera using extended Kalman filter. By using accelerated KAZE features and a more rigorous feature rejection routine combined with 1-point RANSAC for outlier rejection, short baseline matching of features are significantly improved, even with lesser number of feature points, especially in the presence of motion blur. We present a comparative study of our proposal with FAST and show improved localization accuracy in terms of absolute trajectory error.

**Keywords:** EKF · MonoSLAM · AKAZE · Localization

## 1 Introduction

Harris corner detection, SURF or FAST corner detector [8] are the usual feature descriptor of choice while detecting sensible landmarks for localization and mapping. Despite being fast and effective in most situations, they often exhibit poor repeatability in presence of motion blur. While mapping out areas with few corners or flat texture, the system often detects too few landmarks, resulting in poor localization accuracy of the system. We noticed many cases where sudden movement of the camera resulted in a series of motion-blurred frames. In those cases, Harris Corner or FAST does not detect any feature points (beyond an acceptable threshold) and the camera localization becomes significantly erroneous after such maneuvers.

There has been extensive research in MonoSLAM over the last two decades (or more). However, despite the stellar performance of approach like EKF based

MonoSLAM, PTAM [10], DTAM [11] etc., the monocular camera based SLAM paradigm is yet to reach at par with stereo/RGB-D based SLAM frameworks in terms of accuracy.

Moreover, in case of extended Kalman filter based monoslam [6], the feature matching in subsequent frames is done using a normalized cross-correlation of image patches, instead of using a descriptor-to-descriptor comparison across image. This is done to ensure real-time operation of the algorithm. Hence, it is very important to ensure proper initialization of feature points, so that they can be identified easily in the subsequent frames.

### 1.1 Objective

While localizing and mapping a monocular camera using extended Kalman filter, two very specific areas for improvement (over and above the existing state-of-the-art) were identified. The first issue was observed with scenarios where camera exhibits a sudden motion, abruptly changing its pose over a short period of time. The movement induces blurry frames, short baseline matching goes wrong for a few frames and the camera localization suffers considerable loss of accuracy after every such situations. We analyze these conditions and propose an alternative solution for handling this situation better.

A secondary objective is to keep the feature vector size constant while maintaining same accuracy levels. This is done by aggressively pruning the number of feature points being tracked by the filter.

The main contribution of our work is the integration of accelerated KAZE features with EKF based mono SLAM. We show the possibility of obtaining better localization accuracy using AKAZE. We also use 1-point RANSAC for outlier rejection [5] and the combined output has been described in the results.

### 1.2 Related Work

All the filtering based monocular SLAM algorithms work in two recognizable steps: extract features from the image plane and track the features to update the state vector, which typically updates both the camera/robot state as well as the world map. Feature extraction is a key component of this algorithm and considerable research has been done to study the effect of various feature detection techniques on the outcome of the SLAM architecture. [9] compares the effect of SURF, SIFT, BRIEF and BRISK on visual SLAM. [13] proposed ORB as an efficient alternative to SURF and SIFT.

In the following sections, we describe our proposal and compare it with some of the existing techniques.

## 2 Method

We first briefly describe the usual steps associated with the conventional monoslam algorithm based on EKF. Then we present the feature detection mechanism that we incorporated into the process.

The state representation of the pose of the camera is a 13 dimensional vector [1]:

$$x_v = \begin{bmatrix} r^W \\ q^{WC} \\ v^W \\ \omega^C \end{bmatrix} \quad (1)$$

Which can be explained as a 3D position vector $r^W$, unit quaternion $q^{WC}$, velocity vector $v^W$, and angular velocity vector $\omega^C$ relative to a world frame W and a frame C fixed with the camera. Acted upon by an uniform angular and translational velocity, the state transition is formulated by:

$$g_v(\mu_{t-1}) = \begin{bmatrix} r^{WC}_{t-1} + v^W_{t-1}\Delta t \\ q^{WC}_{t-1} \times quat(\omega^C_{t-1}\Delta t) \\ v^W_{t-1} \\ \omega^C_{t-1} \end{bmatrix} \quad (2)$$

where $\mu_{t-1}$ is the previous mean and $\mu_t$ is the current mean. The motion model thus generated is non-linear in nature, since the linear and angular velocity driving the camera is random and cannot be properly predicted.

Given the non-linearity of the state transition, the extended Kalman filter formulation is used for simultaneous state estimation and prediction:

$$\bar{\mu}_t = g(a, \mu_{t-1}) \quad (3)$$

$$\bar{\Sigma}_t = G_t \Sigma_{t-1} G_t^T + R^t \quad (4)$$

$$K_t = \bar{\Sigma}_t H_t^T (H_t \bar{\Sigma}_t H_t^T + Q_t)^{-1} \quad (5)$$

$$\mu_t = \bar{\mu}_t + K_t(z_t - h(\bar{\mu}_t)) \quad (6)$$

$$\Sigma_t = ((\mathbb{1}) - K_t H_t)\bar{\Sigma}_t \quad (7)$$

However, (1) does not represent the entire feature vector. The state space representation used here includes the state of the camera, as well as the entire set of feature points being tracked by the system.

Detecting feature points of interest is a key element of this algorithm. Traditionally, Harris corner detector or Features from Accelerated Segment Test (FAST) are used for detecting key points in an image.

We propose to introduce KAZE features [3] for detecting the landmarks in the image. The scale space is discretized in logarithmic increments and maintained in a series of O octaves and S sub-levels. These indices are mapped to their corresponding scale $\sigma$ by:

$$\sigma_i(o, s) = \sigma_0 2^{\frac{o+s}{S}} \tag{8}$$

The scale space is converted to time units with the mapping:

$$t_i = 1/2\sigma_i^2 \tag{9}$$

Starting from the classic non-linear diffusion formulation:

$$\frac{\partial L}{\partial t} = div(c(x, y, t).\nabla L) \tag{10}$$

where the conductivity c is dependent on the gradient magnitude:

$$c(x, y, t) = g(|\nabla L_\sigma(x, y, t)|) \tag{11}$$

and the function g, as expressed by Perona and Malik [12], can have two different formulation:

$$g_1 = e^{(-\frac{|\nabla L_\sigma|^2}{k^2})}, g_2 = \frac{1}{(1 + \frac{|\nabla L_\sigma|^2}{k^2})} \tag{12}$$

where $k$ is the contrast factor that controls the level of diffusion.

There is no analytical solution for the PDEs involved in Eq. 10, so they are approximated using a semi-implicit scheme. Starting from Eq. 9 and the contrast parameter, the non linear scale space is defined as:

$$L^{i+1} = (I - (t_{i+1} - t_i).\sum_{l=1}^{m} A_l(L^i))^{-1} L^i \tag{13}$$

Over multiple scale levels, the response of scale normalized determinant of Hessian is used for detecting feature points of interest:

$$L_{hessian} = \sigma^2 (L_{xx}L_{yy} - L_{xy}^2) \tag{14}$$

where $L_{xx}$ $L_{yy}$ are the second order horizontal and vertical derivatives respectively. On a set of filtered image Li, a rectangular window of $\sigma^i \times \sigma^i$ is searched for the extrema. Sub-pixel accuracy is not searched for. We also skip the formation of feature descriptor, since the patch matching in subsequent frames will be done by cross correlation of image segments.

To speed up the operation, we use the Fast Explicit Diffusion [2] scheme by performing M cycles of n explicit diffusion steps with non-uniform steps $\tau_j$ that is formed by:

$$\tau_j = \frac{\tau_{max}}{\cos(\pi \frac{2j+1}{4n+2})} \tag{15}$$

where $\tau_max$ is the maximum step that does not violate the stability of the explicit scheme.

The discretization of the diffusion equation can be expressed as:

$$\frac{L^{i+1} - L^i}{\tau} = A(L^i)L^i \tag{16}$$

And given an apriori estimate of $L^{i+1,0} = L^i$, a FED cycle can be expressed as:

$$L^{i+1,j+1} = (I + \tau_j A(L^i))L^{i+1,j} \tag{17}$$

Using this step, we get a feature point (u,v), which needs to be converted to inverse depth parametrization [4]. Basically, the inverse depth parameters is a six dimensional vector, represented by (18):

$$y_i = (x_{c,i}\ y_{c,i}\ z_{c,i}\ \theta_i\ \phi_i\ \rho_i)^T \tag{18}$$

where $x_{c,i}, y_{c,i}, z_{c,i}$ represents the position of the camera w.r.t the world when the feature was first observed, $\theta_i, \phi_i$ represents the azimuth and elevation of the feature point, when observed and $\rho_i$ is the depth estimate of the feature point (which is usually initialized at 0.1).

The feature points, represented in inverse depth, are appended to the camera pose vector to form the state vector of the system. This vector is iteratively predicted and measured by EKF Eqs. (3) through (7).

The rest of the EKF measurement and update is done by standard formulation, with two step partial update for low and high innovation inliers in a RANSAC hypothesis [5].

Moreover, we do not allow any feature's inverse depth parameters to persist in the feature vector beyond 3 cycles of EKF, thereby reducing the rate of increase of feature vector size.

## 3 Results

We use Absolute Trajectory Error (ATE) as a means to validate our approach. ATE compares the trajectory of a robot/camera, as reconstructed by an algorithm using real sensor data as its input, to the actual trajectory (ground truth).

We benchmark our approach in the RGB-D SLAM dataset of TUM [14,15]. Only RGB data is used for the experiments, while the groundtruth trajectory provided in the dataset is used for validation. The EKF is implemented on MATLAB (which is based on the open source code provided by [10]) while the computation of AKAZE features is done in C++. We observed approximately 20–25 % decrease in root mean squared error of absolute trajectory over short sequences, using the technique we proposed in the previous section (Table 1).

The dataset we have used for the demonstration example is *freiburg1_room* from [14]. The image sequence has been captured using a Microsoft Kinect. The experiment where we use FAST feature descriptor along with existing filter based monoslam algorithm has been denoted **EXP A** (Fig. 1). Our proposed approach has been denoted **EXP B** (Figs. 2 and 3).

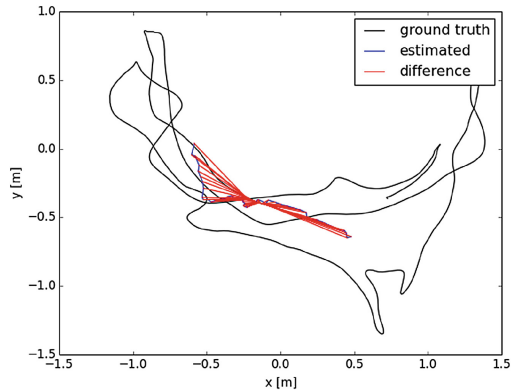
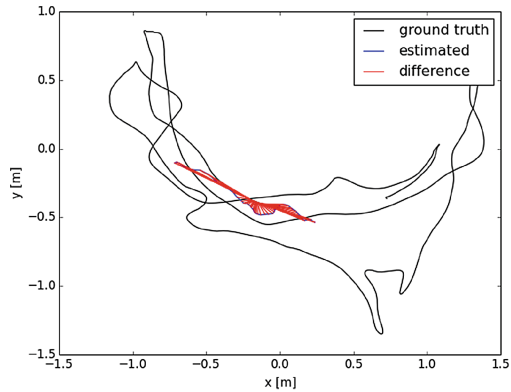

**Fig. 1.** Localization using FAST

**Fig. 2.** Localization using AKAZE

**Table 1.** Comparison of results between EXP A (FAST) and EXP B (AKAZE)

|  | EXP A | EXP B |
|---|---|---|
| Root Mean Square Error | 0.320698 m | 0.243540 m |
| Mean | 0.278879 m | 0.206980 m |
| Median | 0.232776 m | 0.155320 m |
| Standard Deviation | 0.158348 m | 0.128339 m |
| Min. Error | 0.092825 m | 0.086545 m |
| Max. Error | 0.619539 m | 0.561998 m |

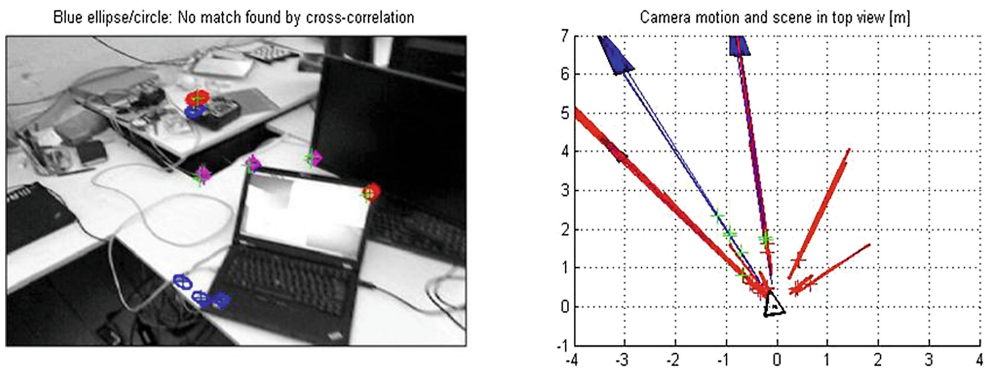

**Fig. 3.** The circles represent the feature points detected by AKAZE. The red ellipses are the matched points, the pink ones are those rejected by 1-point RANSAC (Color figure online)

**Table 2.** Comparison of results between EXP C (ORB) and EXP D (AKAZE)

|  | EXP C | EXP D |
|---|---|---|
| Root Mean Square Error | 1.213417 m | 1.150064 m |
| Mean | 1.073303 m | 1.023851 m |
| Median | 1.160499 m | 1.109093 m |
| Standard Deviation | 0.566041 m | 0.523809 m |
| Min. Error | 0.171942 m | 0.103387 m |
| Max. Error | 2.306686 m | 2.084978 m |

For the sake of completeness, we also compared the proposed approach with feature detection using ORB. The results obtained are tabulated below. **EXP C** denotes the results obtained while using ORB as the feature detector. **EXP D** denotes the proposed framework using AKAZE. This was done on the data set *freiburg1_360*, which proved to be more error prone (in terms of short-baseline localization accuracy) due to the presence of heavy motion blur. Even in this experiment, AKAZE performed better than ORB. However, the advantage was slightly less pronounced (Table 2).

The time performance of AKAZE is better than SURF or SIFT, but not as efficient as FAST [7]. The extended Kalman filter based mono SLAM section of the proposed algorithm is mostly similar to [5].

## 4 Conclusion

Using accelerated KAZE features for feature point detection in monoslam is not documented in any of the literature we surveyed so far. It results in better localization accuracy in dataset involving motion – blurred frames. This has been validated in RGB-D dataset by comparison against ground truth values. MonoSLAM is a field of study which has immense scope for improvement in terms of accuracy and reliability. It is necessary to benchmark the performance of MonoSLAM using various feature detectors. Although both the original MonoSLAM algorithm and AKAZE runs in real time, this research work does not cover the time performance of the two combined. This needs to be analyzed further.